\documentclass[11pt]{article}

\usepackage[final]{acl}

\usepackage{times}
\usepackage{latexsym}

\usepackage[T1]{fontenc}

\usepackage[utf8]{inputenc}

\usepackage{microtype}

\usepackage{inconsolata}

\usepackage{graphicx}

\usepackage{algorithm}
\usepackage{algorithmic}
\usepackage{amsmath}
\usepackage[table]{xcolor}
\definecolor{mygray}{rgb}{0.8,0.8,0.8}
\definecolor{myorange}{rgb}{0.8, 0.35, 0.0}
\usepackage{makecell}
\usepackage{booktabs}
\usepackage{multirow}
\usepackage{subfig}
\definecolor{deepblue}{HTML}{0000FF} 
\usepackage{tcolorbox}
\newcommand{\up}[1]{\textcolor{red}{{$\uparrow$#1}}}
\newcommand{\down}[1]{\textcolor{red}{{$\downarrow$#1}}}
\definecolor{tabcolor3}{rgb}{0.86, 0.93, 1.0}
\newcommand{\vv}{\mathbf{v}}

\newcommand{\ours}{SCS }
\usepackage{amssymb,mathrsfs}
\newcommand{\paren}[1]{\left(#1\right)}
\usepackage{wrapfig} 

\title{Focusing Condition: Inference-Time Self-Contrastive Steering Elicits\\ Better Conditional Text Embeddings in LLMs}

\author{
Zifeng Cheng$^\text{\textnormal{1}}$ \quad \quad
Lingyun Qian$^\text{\textnormal{1}}$ \quad \quad
Zhiwei Jiang$^\text{\textnormal{1}}$\thanks{Corresponding author.} \quad \quad \\
\textbf{
Cong Wang$^\text{\textnormal{1}}$ \quad \quad
Yafeng Yin$^\text{\textnormal{1}}$ \quad \quad
Fei Shen$^\text{\textnormal{2}}$ \quad \quad 
Ao Zhou$^\text{\textnormal{1}}$ \quad \quad
Qing Gu$^\text{\textnormal{1}}$
} \\
$^\text{1}$ State Key Laboratory for Novel Software Technology, Nanjing University \\
$^\text{2}$ National University of Singapore
\\
\texttt{chengzf@nju.edu.cn} \quad 
\texttt{522024330064@smail.nju.edu.cn} \quad \\
\texttt{\{jzw,wang.c,yafeng\}@nju.edu.cn} \quad
\texttt{shenfei29@nus.edu.sg} \\
\texttt{za@smail.nju.edu.cn} \quad 
\texttt{guq@nju.edu.cn}
}

\begin{document}
\maketitle
\begin{abstract}
Extracting conditional text embeddings from large language models (LLMs) is a promising paradigm, as it requires neither additional data nor fine-tuning.
Existing methods incorporate conditions into prompts to guide LLMs to focus on specific aspects and elicit conditional text embeddings.
However, relying solely on prompts often fails to produce high-quality conditional text embeddings, as they remain entangled with general text embeddings, ultimately degrading their quality.
To this end, we propose an inference-time, plug-and-play Self-Contrastive Steering (SCS) method that constructs unconditional general text embeddings and uses them to refine conditional text embeddings, making them more focused on the target condition.
Specifically, we modify the attention mask and positional encodings to mask the condition, thereby obtaining unconditional text embeddings and intervening in the multi-head self-attention computation process.
Notably, our method is highly efficient, requiring only a single additional multi-head self-attention computation at inference time.
Extensive experiments on clustering, Semantic Textual Similarity, and triplet alignment datasets demonstrate that our method can seamlessly improve the performance of existing prompt-based methods across different LLMs in a training-free and plug-and-play manner.
Our code will be released at \url{https://github.com/zifengcheng/SCS}.
\end{abstract}

\section{Introduction}
\begin{figure}[t]
    \centering
    \includegraphics[width=0.48\textwidth]{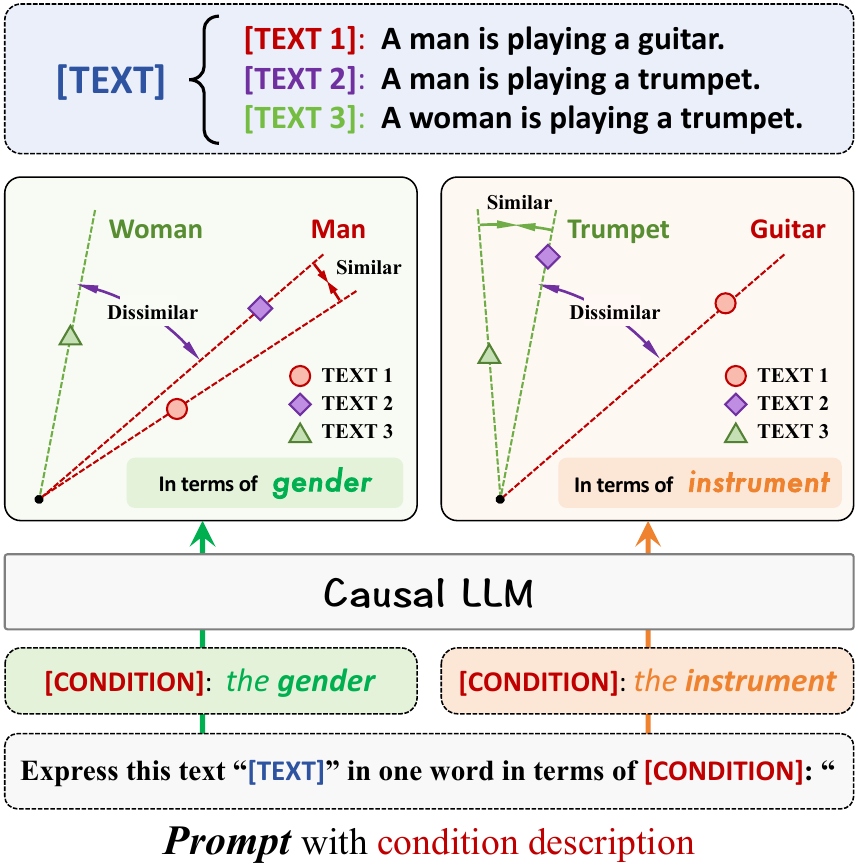}
    \caption{An example of the conditional text embedding task. Notably, compared to labels, conditions often correspond to coarser-grained information.
    }
    \label{fig:moti}
\end{figure}

Text embeddings aim to represent a text as a generic fixed-size vector and have a wide range of applications, including text classification, clustering, semantic textual similarity~\cite{sts12,sts13}, and information retrieval.
In practice, text often contains multiple types of semantics, and a single universal embedding cannot satisfy users' specific fine-grained embedding needs from multiple aspects~\cite{csts,DBLP:conf/emnlp/WangSZ23}.
For example, as shown in Figure~\ref{fig:moti}, TEXT 1 and TEXT 2 are similar with respect to \textit{gender}, but they differ with respect to the \textit{instrument}.
Therefore, conditional text embeddings are crucial for a wide range of downstream tasks.

Given the success of Large Language Models (LLMs) on a wide range of NLP tasks in zero-shot settings, several studies \cite{ponte,prompteol,DBLP:conf/iclr/SpringerKFNR25,lei2024meta} focus on \textit{prompt engineering} to directly extract text embeddings from LLMs \textit{without any data and further training}.
This training-free, zero-shot setting offers a substantial improvement over methods that rely on extensive fine-tuning to achieve high performance.
For example, PonTE \cite{ponte} builds on PromptEOL~\cite{prompteol} by adding condition and introduces a simple and effective \textit{semantic compression} prompt that compresses the conditional text embedding into the last token: \textit{Express this text:} ``\texttt{[TEXT]}'' \textit{in one word in terms of ``\texttt{[CONDITION]}'': ``}, where \texttt{[TEXT]} and \texttt{[CONDITION]} serve as placeholders for the text and condition.
The phrase ``in one word'' encourages the LLM to condense the semantic meaning of the text into the last token.

Even if these methods are effective, embeddings obtained by relying solely on the conditions in the prompt still encode other condition-irrelevant information.
Empirically, we observe that conditional text embeddings still encode general textual information, which affects their distance measurements and ultimately undermines their performance.
For example, the average cosine similarity between general text embeddings and conditional text embeddings reaches as high as 0.763 on the TweetEmotion dataset.
Moreover, prompts are sensitive to subtle variations~\cite{DBLP:conf/acl/LuBM0S22}, and the mechanisms underlying their effectiveness remain unclear~\cite{DBLP:conf/icml/ShiWXL24}.

To solve these issues, we propose a simple and effective plug-and-play Self-Contrastive Steering (SCS) method to enhance existing prompt engineering methods.
SCS adjusts the attention mask and positional encoding to derive multi-view embeddings from the text, including both unconditional text embeddings and conditional text embeddings.
By contrasting the two embeddings, SCS can steer the embeddings to focus on condition-related information while filtering out unconditional general text information.
Specifically, we first forward propagate the prompt-wrapped text to the specific layer and obtain the hidden states.
Next, we obtain an unconditional text embedding by modifying the attention mask and positional encoding, allowing us to compare it with the conditional text embedding.
Furthermore, we propose two self-contrastive activation steering methods, direct elimination and projection-based elimination, to purify the conditional text embeddings.
Finally, we continue to forward propagate the refined representations to obtain the conditional text embeddings.

Our main contributions are as follows:
\begin{itemize}
\item We first propose a lightweight approach to create multi-view embeddings for a text by modifying the attention mask and positional encoding.
\item We propose a Self-Contrastive Steering method, which leverages two self-generated embeddings to guide the model to focus on the conditional part.
\item We conduct extensive experiments on seven datasets spanning clustering, Semantic Textual Similarity (STS), and triplet alignment tasks. Experimental results demonstrate that our proposed method significantly improves the performance of existing prompt-based methods across different LLMs.
\end{itemize}

\section{Related Work}
\paragraph{Text Embedding}
Text embedding aims to use a general fixed-size vector to represent a text, playing a fundamental role in various applications \cite{DBLP:conf/aaai/YangCJYWGFG26}.
Existing methods typically use contrastive learning to fine-tune models for text embedding~\cite{gao2021simcse,DBLP:conf/emnlp/JiangJHZWZWHDZ22,DBLP:conf/acl/NiACMHCY22}, where positive pairs are pulled closer and negative pairs are pushed apart.
A classic work is SimCSE~\cite{gao2021simcse}, which uses dropout to generate positive samples for contrastive learning.
Recently, some works~\cite{li2024bellm,behnamghader2024llm2vec} have begun to fine-tune LLMs to obtain text embeddings.

\paragraph{Conditional Text Embedding}
Unlike general text embeddings, conditional text embeddings can selectively focus on specific aspects of a text to more precisely meet user needs at a finer granularity.
Existing approaches primarily fine-tune LLMs to obtain conditional text embedding.
InstructOR \cite{DBLP:conf/acl/SuSKWHOYSZ023} first integrates instructions into text embeddings for generating task-relevant embeddings.
It annotates instructions for 330 different tasks and employs contrastive learning for training.
C-STS \cite{csts} first annotates a conditional semantic textual similarity dataset to accurately evaluate the similarity between sample pairs.
InBedder \cite{InBedder} leverages the generative capabilities of LLMs to re-encode the generated content into embeddings and fine-tunes an LLM on preprocessed QA datasets with short responses.
HyperCL \cite{DBLP:conf/acl/YooCKK24} integrates hypernetworks with contrastive learning to compute conditioned sentence representations.
However, these methods often require large amounts of data and incur high fine-tuning costs, while also sacrificing the general capabilities of the LLM by turning it into an embedding model.

\paragraph{Extracting Conditional Embeddings from LLMs}
Early work primarily focus on extracting general text embeddings from LLMs using prompt engineering methods~\cite{prompteol,DBLP:conf/iclr/SpringerKFNR25,lei2024meta}.
For example, PromptEOL~\cite{prompteol} first proposes a simple and effective \textit{semantic compression} prompt that compresses the text embedding into the last token: \textit{This sentence:} ``\texttt{[TEXT]}'' \textit{means in one word: ``}.
Subsequently, considering the causal attention property of LLMs, Echo~\cite{DBLP:conf/iclr/SpringerKFNR25} proposes duplicating the text twice, while TP~\cite{DBLP:conf/acl/FuCJW0L025} prepends the sentence embedding token to allow all tokens in the sentence to perceive the complete semantics.
Recently, considering the need for more fine-grained embeddings, PonTE~\cite{ponte} first adds conditions on top of PromptEOL to extract more fine-grained conditional text embeddings.
However, relying solely on the conditions in the prompt cannot ensure that the output embeddings capture only condition-relevant information.

\begin{figure*}[t]
    \centering
    \includegraphics[width=1\textwidth]{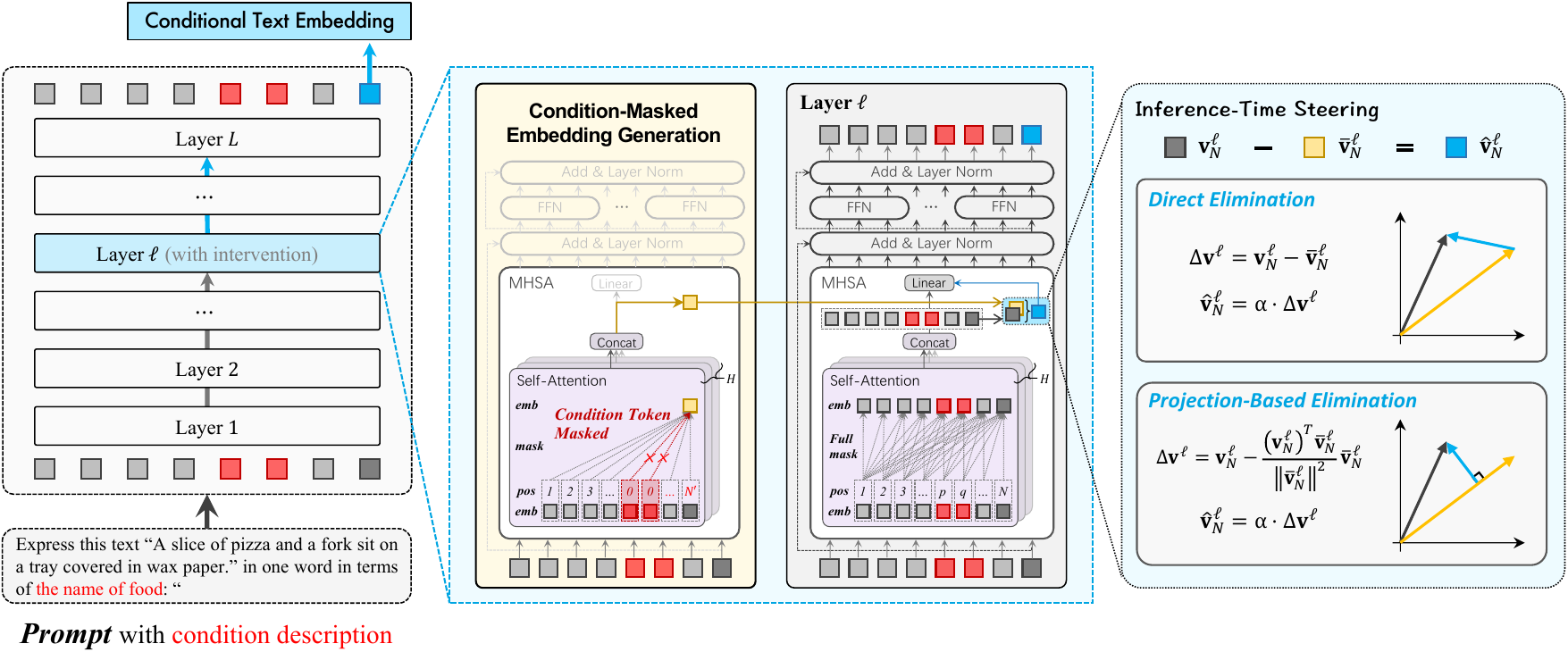}
    \caption{Illustration of the self-contrastive steering method. Our method additionally generates an unconditional text embedding through condition-masked embedding generation and introduces two activation steering strategies.}
    \label{fig:framework}
\end{figure*}

\paragraph{Activation Steering}
Activation steering~\cite{RE} is a training-free technique that creates steering vectors to modify the activations of LLMs, thereby controlling their generation.
The first type of method~\cite{DBLP:conf/acl/RimskyGSTHT24,DBLP:conf/nips/0002PVPW23,chengsteering} generates steering vectors from positive-negative sample pairs, often relying on supervised data.
The second type~\cite{DBLP:conf/emnlp/LeongCWWL23,postmus,DBLP:conf/acl/ChengWFJ00025} constructs auxiliary prompts through forward passes, typically incurring additional computational overhead.
Unlike these methods, our approach requires no supervised data, constructs input-related steering vectors, and only requires a single additional multi-head self-attention computation.

\section{Preliminary}
\paragraph{Extracting Conditional Text Embeddings from LLMs}
Existing methods mainly focus on prompt engineering techniques to \textit{compress} the semantics of a text into a single token for extracting embeddings from LLMs.
PonTE~\cite{ponte} builds upon PromptEOL~\cite{prompteol} by incorporating additional conditions to extract conditional text embeddings:
\begin{tcolorbox}
\textbf{PonTE}: Express this text ``\texttt{[TEXT]}'' in one word in terms of \texttt{[Condition]}: ``
\end{tcolorbox} 
where \texttt{[TEXT]} and \texttt{[Condition]} denote placeholders for the text and the condition, respectively.
The phrase ``in one word'' encourages the LLM to condense the semantic meaning of the text into the hidden state of the next token.
Notably, the hidden state of the last token `` already encodes information about the next token for next-token prediction, allowing us to directly use its representation as the conditional text embedding without generation.

\paragraph{Self-Attention with RoPE in LLMs}
LLMs employ Rotary Position Embedding (RoPE) \cite{rope} in the Multi-Head Self-Attention (MHSA) module to encode relative positional information by applying a position-dependent rotation to each query and key vector before computing attention.
Specifically, the $\ell$-th MHSA layer receives the hidden state at position $m$ from the previous layer and transforms it into the corresponding query and key representations as follows:
\begin{equation}
\mathbf{q}^{\ell}_m = \text{RoPE}_q(\mathbf{h}^{\ell-1}, m) = \mathbf{W}_q^{\ell} \mathbf{h}^{\ell-1} e^{i m \theta}
\end{equation}
\begin{equation}
\mathbf{k}^{\ell}_m = \text{RoPE}_k(\mathbf{h}^{\ell-1}, m) = \mathbf{W}_k^{\ell} \mathbf{h}^{\ell-1} e^{i m \theta}
\end{equation}
where $\mathbf{W}_q^{\ell}$ and $\mathbf{W}_k^{\ell}$ denote the query and key projection matrices at layer $\ell$, and $\theta$ is a dimension-dependent constant that determines the rotation frequency.

\section{Method}
Our proposed \ours method is an inference-time plug-and-play steering method that requires no additional data or fine-tuning of the LLM, and can integrate with existing prompt engineering techniques.

\ours consists of three steps to obtain conditional text embedding, as illustrated in Figure~\ref{fig:framework}.
In the first step, it inputs the prompt-wrapped input into the LLM up to the $(\ell-1)$-th Transformer layer to obtain the output hidden states.
Next, it modifies the attention mask to hide the conditional information in the prompt and further adjusts the positional encodings to make them contiguous, thereby extracting an unconditional text embedding at layer $\ell$.
Finally, it performs self-contrast between the two embeddings obtained with and without the condition to remove general information from the conditional text embedding, and then continues to forward the purified representations to obtain the final conditional text embedding.

\subsection{Prompt Construction}
In the first step, we construct a prompt (i.e., Express this text ``\texttt{[TEXT]}'' in one word in terms of ``\texttt{[CONDITION]}'': ``) to wrap the text and feed it into the LLM to obtain the hidden states from the $(\ell-1)$-th layer.

Formally, given a prompt-wrapped input $T = [t_1, \ldots, t_{p-1}, t_p, \ldots, t_q, t_{q+1}, \dots, t_N]$, where $t_p$ to $t_q$ constitute the conditional part and $N$ denotes the input length, we feed it into the LLM and obtain the output hidden states (i.e., $\mathbf{H}^{\ell-1} = [\mathbf{h}^{\ell-1}_1, \cdots, \mathbf{h}^{\ell-1}_N]$) at the $(\ell-1)$-th layer.

\subsection{Multi-View Embedding Extraction}
In the second step, we extract multi-view text embeddings, including the conditional text embedding and the unconditional text embedding.
For the \textit{conditional text embedding}, we perform standard MHSA computation without modifying the attention mask or positional encoding to obtain the contextualized value vectors (i.e., $\mathbf{V}^{\ell} = [\vv^{\ell}_1, \cdots, \vv^{\ell}_{N}]$);
for the \textit{unconditional text embedding}, we modify the attention mask and positional encoding in the MHSA module to mask out the conditional part of the input.
Then, we describe the technical details of extracting the unconditional text embedding.

\textbf{Attention Mask} We first modify the attention mask to prevent the tokens following the condition from seeing the condition, thereby extracting the unconditional text embedding.
Since we only focus on the last token for the unconditional text embedding, the attention mask for the last token can be defined as follows:
\[
\mathbf{m}^{\ell}_N(i) =
\begin{cases}
0, & 1 \leq i \leq p-1, \\
-\infty, & p \leq i \leq q, \\
0, & q+1 \leq i \leq N .
\end{cases}
\]
where $\mathbf{m}^{\ell}_N(i)$ denotes the attention mask value of the $N$-th token to the $i$-th token in the $\ell$-th layer.

\textbf{Positional Encoding} After applying the attention mask, the positional encodings of the text become discontinuous.
Specifically, there are no tokens with positional encodings from $p$ to $q$, which can affect the quality of the unconditional text embeddings.

Thus, we further adjust the positional encoding to align with the modified attention mask.
Specifically, we shift the positional encodings of all tokens following the condition forward by q-p+1 positions to ensure positional continuity.
The new positional encoding function pos($i$) is defined as follows:
\[
\text{pos}(i) =
\begin{cases}
i, & 1 \le i \le p-1, \\
0, & p \le i \le q, \\
i - (q-p+1), & q+1 \le i \le N.
\end{cases}
\]
Since the condition part (i.e., $p \le i \le q$) is masked, its positional encoding settings do not affect the final results.
We simply set them to 0.
Notably, after modifying the attention mask and positional encodings, the resulting new prompt forms a coherent PromptEOL-like prompt, which can be used to extract high-quality general unconditional text embeddings.

\textbf{Self-Attention Computation}
After defining the attention mask and positional encoding, we incorporate positional information into the query and key vectors using RoPE before computing the attention scores:
\[
\alpha^{\ell}_{N,i} 
= 
\text{softmax}_i(\frac{
    (\mathbf{q}^{\ell}_{\text{pos($N$)}})^{\!\top}
    \mathbf{k}^{\ell}_{\text{pos($i$)}}
}{\sqrt{d}}
+ \mathbf{m}^{\ell}_N(i))
\]
where $\alpha^{\ell}_{N,i}$ denotes the attention score from the $N$-th token to the $i$-th token and $\mathbf{q}^{\ell}_{\text{pos($N$)}}$ represents the query vector after injecting the new positional encoding function.

Then, we can aggregate the value vectors of all tokens to obtain the unconditional text embedding: 
\begin{equation}
    \bar{\vv}^{\ell}_N = \sum_{i=1}^{N} \alpha^{\ell}_{N,i} \paren{\mathbf{W}^{\ell}_V \mathbf{h}^{\ell-1}_i}
\end{equation}
where $\bar{\vv}^{\ell}_N$ represents the contextualized value vectors used to encode the unconditional text embedding and $\mathbf{W}^{\ell}_V$ denotes the value projection matrix in the $\ell$-th MHSA layer.

Notably, we define the above process on a single head, and it can be extended to the outputs of multiple heads in MHSA.
The overhead of unconditional embedding extraction is very light, as it only requires one additional forward pass through the MHSA layer.

\subsection{Self-Contrastive Activation Steering}
In the final step, it contrasts the two embeddings obtained from itself to purify the conditional embedding, and then continues forward propagation to obtain the final conditional embedding.
We propose two alternative self-contrastive steering strategies to obtain the conditional steering vector: direct elimination and projection-based elimination.

\textbf{Direct Elimination (DE)} contrasts two embeddings to obtain the conditional steering vector $\Delta\mathbf{v}^{\ell}$.
Intuitively, these two embeddings differ only in the conditional part, and comparing them can highlight the conditional information in the prompt.
Specifically, this vector is calculated as:
\begin{align}
    \Delta\vv^{\ell}_N = \vv^{\ell}_N - \bar{\vv}^{\ell}_N
    \label{eq:semantic_vector}
\end{align}

\textbf{Projection-Based Elimination (PE)} first projects the conditional text embedding onto the unconditional text embedding to identify the condition-independent components, and then eliminates them through a contrastive operation.
Specifically,
\begin{equation}
\Delta\vv^{\ell}_N = \vv^{\ell}_N - \frac{(\vv^{\ell}_N)^{\!\top} \bar{\vv}^{\ell}_N}{\|\bar{\vv}^{\ell}_N\|^2} \bar{\vv}^{\ell}_N
\end{equation}
Notably, the conditional steering vector is text-specific, with each text having its own distinct steering vector.
In addition, we can also apply steering on the hidden states or the outputs of the FFN.

Since the norms of the vector change significantly before and after the contrast, we introduce a scaling factor $\alpha$ to control the norm of the vector after the intervention.
\begin{equation}
\hat{\vv}^{\ell}_{N} =  \alpha \cdot \Delta\vv^{\ell}_N
\end{equation}

After getting refined conditional embedding $\hat{\vv}^{\ell}_{N}$, we further use $\hat{\vv}^{\ell}_{N}$ to replace original conditional text embedding $\vv^{\ell}_{N}$ and obtain the new input $\hat{\mathbf{V}}^{\ell}$ for the output matrix in the MHSA layer.
Specifically,
\begin{align}
   \hat{\mathbf{V}}^{\ell} = [\vv^{\ell}_1, \cdots, \hat{\vv}^{\ell}_{N}].
\end{align}

Finally, we feed the refined contextualized value vector $\hat{\mathbf{V}}^{\ell}$ into the output matrix $\mathbf{W}^{\ell}_O$ of the $\ell$-th MHSA layer and continue the standard forward propagation to extract conditional text embeddings.

\begin{table*}[t]
\centering
\resizebox{\linewidth}{!}{%
\begin{tabular}{l ccc cc cc c}
  \toprule
  \multirow{2.5}{*}{\textbf{Model}}
  & \multicolumn{3}{c}{\textbf{Clustering (V-measure $\uparrow$)}} & \multicolumn{2}{c}{\textbf{STS (Spearman \& Pearson $\uparrow$)}} & \multicolumn{2}{c}{\textbf{Triplet Alignment (Acc. $\uparrow$)}} & \\ 
  \cmidrule(lr){2-4} \cmidrule(lr){5-6} \cmidrule(lr){7-8}
   & \textbf{TweetEmo} & \textbf{Amazon-R} & \textbf{Amazon-C} & \textbf{I.STSB} & \textbf{C-STS} & \textbf{Toxic} & \textbf{AGNews} & \multirow{-2.5}{*}{\textbf{Mean $\uparrow$}} \\ 
  \midrule
  \midrule
  \multicolumn{8}{l}{\it{Supervised Contrastive Training}}\\
  \textbf{sup-SimCSE$_{\text{RoBERTa-large}}$} & 29.4 & 22.4 & 19.5 & - & 4.1/3.4 & - & - & - \\
  \textbf{E5$_{\text{Mistral-7B-Inst}}$} & 41.3 & 37.6 & 37.4 & - & 34.8/34.6 & - & - & -\\
  \textbf{GTE$_{\text{Qwen2-7B-Inst}}$} & 36.8 & 36.8 & 38.3 & - & 33.5/33.9 & - & - & - \\
  \midrule
  \midrule
  \multicolumn{8}{l}{\it{Without Contrastive Training}}\\
  \textbf{PromptEOL$^\dag$}  & 31.7 & 30.8 & 9.4 & 0/0  &3.2/2.5  &55.0 &71.5 &22.7 \\
  \textbf{PonTE$^\dag$} & 39.9 & 34.8 & 32.5 & 33.9/29.5 & 36.2/32.4 & 57.7 & 84.2 &42.3\\
  \textbf{PonTE + TP$^\dag$} &42.6 &35.4 &33.6 &30.7/25.5 &34.6/31.4 &56.9 &82.5 &41.4\\
  \textbf{PonTE + Att$^\dag$} & 43.4 & 32.4 & 33.2 & 33.3/29.1 & 36.8/32.1 & 57.5 & \textbf{85.8} &42.4\\
  \textbf{PonTE + Echo$^\dag$} &42.0 &33.8 &33.5 &33.0/28.6 &35.0/30.8 &57.8 &83.1 &42.0\\
  \textbf{PonTE + SCS-DE (\textbf{\textit{Ours}})} 
& \cellcolor{tabcolor3} \underline{45.5} \up{5.6}
& \cellcolor{tabcolor3} \underline{37.0} \up{2.2}
& \cellcolor{tabcolor3} \textbf{34.2} \up{1.7}
& \cellcolor{tabcolor3} \textbf{39.9/37.7} \up{6.0/8.2}
& \cellcolor{tabcolor3} \textbf{37.9/}\underline{34.5} \up{1.7/2.1}
& \cellcolor{tabcolor3} \textbf{58.0} \up{0.3}
& \cellcolor{tabcolor3} \underline{85.6} \up{1.4}
& \cellcolor{tabcolor3} \textbf{45.6} \up{3.3}\\

  \textbf{PonTE + SCS-PE (\textbf{\textit{Ours}})} 
& \cellcolor{tabcolor3} \textbf{45.9} \up{6.0} 
& \cellcolor{tabcolor3} \textbf{37.1} \up{2.3}
& \cellcolor{tabcolor3} \underline{33.9} \up{1.4}
& \cellcolor{tabcolor3} \underline{38.6/36.0} \up{4.7/6.5}
& \cellcolor{tabcolor3} \textbf{37.9/34.6} \up{1.7/2.2}
& \cellcolor{tabcolor3} \underline{57.9} \up{0.2}
& \cellcolor{tabcolor3} \textbf{85.8} \up{1.6}
& \cellcolor{tabcolor3} \underline{45.3} \up{3.0}\\
  \bottomrule
\end{tabular}
}
\caption{Results on three downstream tasks across seven datasets using LLaMA3-8B-Inst. Since PonTE does not open-source code, $\dag$ denotes our reproduced results. The best results are in \textbf{bold}, and the second-best are \underline{underlined}.}
\label{tab:results}
\end{table*}

\section{Experiments}
\subsection{Datasets and Experimental Settings}
We evaluate conditional text embeddings on the clustering, STS, and triplet alignment tasks.
The clustering task includes 3 datasets.
The Tweet emotion intensity dataset (\textbf{TweetEmo})~\cite{TweetEmo} consists of tweets annotated with corresponding emotion labels.
\textbf{Amazon-R} and \textbf{Amazon-C} are constructed from the Amazon English review corpus.
Amazon-R uses star ratings as labels, while Amazon-C uses product categories as labels.
The STS task includes 2 datasets.
InstructSTSB (\textbf{I.STSB})~\cite{InBedder} extends the original STS-B dataset by incorporating conditions.
\textbf{C-STS}~\cite{csts} is a human-annotated dataset designed to assess the semantic similarity between two sentences under specific conditions.
The semantic similarity scores of sentence pairs in the C-STS dataset range from [0, 5], while those in the InstructSTSB dataset range from [0, 1].
The triplet alignment task includes 2 datasets: Toxic and AGNews.
For each dataset, we randomly sample 10,000 triplets, where the anchor and the positive share the same label, while the negative has a different label.
\textbf{Toxic}~\cite{toxic} consists of comments from the Civil Comments platform annotated for toxicity, while \textbf{AGNews}~\cite{AGNews} is an English news classification corpus containing approximately 127,600 samples across four categories.


For the clustering task, we use K-means for clustering and V-measure~\cite{V-Measure} to evaluate, with the number of clusters provided in advance. 
The final result is the average over five runs.
For the STS task, we use Spearman and Pearson correlation coefficients.
For the triplet alignment task, we report triplet alignment accuracy, defined as the proportion of triplets for which the relative ordering is correctly captured.

We use grid search to search for the intervention layer $\ell$ in [2, 20] and the scaling factor of the norm scaling $\alpha$ in \{5, 10, 15\} for each dataset.
The detailed conditions for clustering and triplet alignment datasets are provided in the Appendix~\ref{app:prompt}, and the STS datasets include these conditions.


\subsection{Baselines}
Our method is compared with two groups of baseline methods. 
The first category is supervised learning methods.
\textbf{sup-SimCSE$_{\text{RoBERTa-large}}$}~\cite{gao2021simcse} uses NLI datasets and contrastive loss to fine-tune the model.
\textbf{E5$_{\text{Mistral-7B-Inst}}$}~\cite{E5} is a two-stage training method: it first performs unsupervised contrastive learning on large-scale datasets, and then fine-tunes on supervised datasets.
\textbf{GTE$_{\text{Qwen2-7B-Inst}}$} \cite{GTE} further expands the amount of data used in the two stages.
The second category is training-free methods.
\textbf{PromptEOL} \cite{prompteol} is an unconditional prompt-based method that compresses the full semantics of the text into the last token.
\textbf{PonTE} \cite{ponte} further incorporates conditions into the prompt to steer the LLM's focus toward specific aspects.
\textbf{PonTE + TP}~\cite{DBLP:conf/acl/FuCJW0L025} prepends the text embedding token to allow each token in the text to capture the full semantics.
\textbf{PonTE + Echo} \cite{DBLP:conf/iclr/SpringerKFNR25} duplicates the text twice to enable the subsequent tokens to capture the complete semantics of the text.
\textbf{PonTE + Att} introduces an additional hyperparameter to amplify the attention scores of the tokens following the condition toward the condition.
The detailed prompts are shown in the Appendix~\ref{app:baseline}.

\begin{figure*}[t]
    \centering
    \includegraphics[width=1\textwidth]{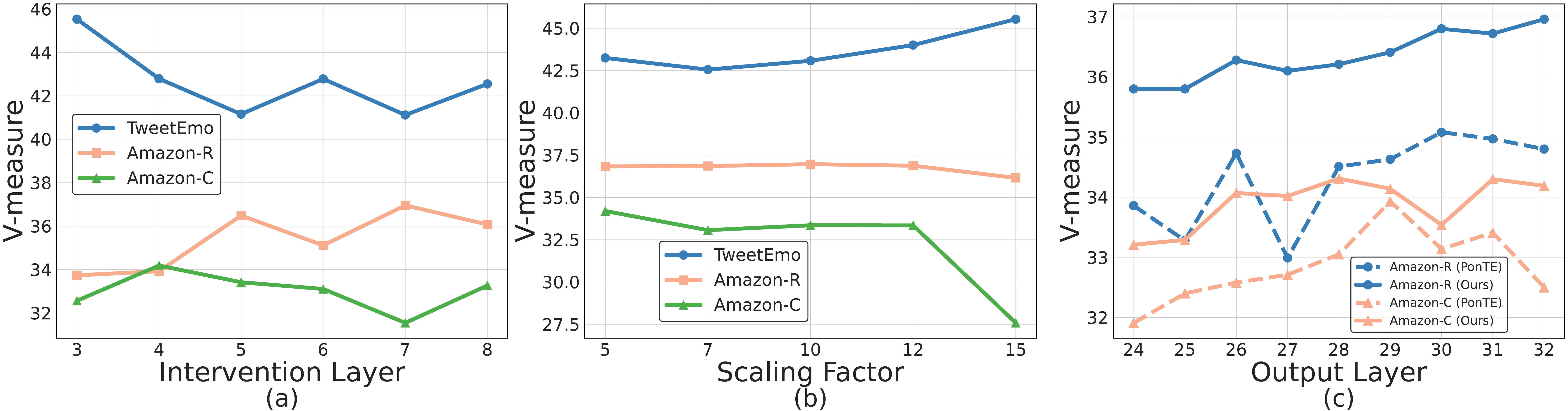}
    \caption{Effects of intervention layer, scaling factor, and output layer on SCS-DE. (a) Effects of the intervention layer. (b) Effects of the scaling factor. (c) The effects of the output layer.}
    \label{fig:layer_analysis}
\end{figure*}

\subsection{Results}
The results of our method on the three tasks are shown in Table \ref{tab:results}.
Our method achieves the best performance in 6 out of 7 cases and achieves the optimal average performance, confirming the effectiveness of our approach.
Our method achieves improvements across all seven datasets.
Notably, SCS-DE achieves improvements of 6\% and 8.2\% on the two metrics of the I.STSB dataset.
This is because the data in this dataset are shorter and the similarity labels are divided with finer granularity.
Furthermore, the even simpler SCS-DE method achieves better average performance, indicating that the projection may be unnecessary.
Besides, our method achieves performance comparable to supervised contrastive learning approaches, demonstrating the potential of directly extracting conditional text embeddings from LLMs.

For the baselines, the unconditional PromptEOL method performs poorly due to the lack of conditions, while PonTE significantly improves PromptEOL's performance by incorporating conditions.
PonTE + Att and PonTE + Echo have minimal impact on the final results, whereas PonTE+TP causes a significant decline.
This may be because instruction-tuned models struggle to comprehend the abstract semantics of the prepended token.

\subsection{Analysis of Additional Inference Time Overhead}
We further compare the time overhead of our method with the baseline methods using a V100 GPU and the same batch size in Table \ref{tab:time}.
We can see that our method introduces almost no additional time overhead, with the fluctuations mainly due to randomness.
This is because our method only computes an additional value vector in a single MHSA layer, while the main computational overhead in LLMs comes from the feed-forward networks.

\begin{table}[t] \small
    \centering
    \setlength{\tabcolsep}{6pt}
    \begin{tabular}{lcc}
        \toprule
        \textbf{Method} & \textbf{TweetEmo} & \textbf{Amazon-R}\\
        \midrule
        \textbf{PonTE}  & 70ms   & 79ms \\
        \textbf{PonTE + SCS-DE (\textbf{\textit{Ours}})} & 69ms &  80ms\\
        \bottomrule
    \end{tabular}
    \caption{The runtime per sample on two datasets using LLaMA3-8B-Inst.}
    \label{tab:time}
\end{table}

\subsection{Model Analysis}
Since SCS-DE is simpler and more effective, the subsequent analysis mainly focuses on it.


\textbf{Effects of the Intervention Position}
We first explore the effects of intervention position on three datasets using LLaMA-3-8B-Inst.
Intervening at all three positions can effectively improve performance, indicating the effectiveness of our method, as shown in Table~\ref{tab:position}.
Notably, the optimal intervention positions vary across datasets, which may be determined by the characteristics of each dataset.
The FFN and hidden variants incur theoretically higher time overhead, as they need to generate unconditional information within the MHSA module and continue forward propagation for steering.
In particular, the hidden variant needs to run through a complete Transformer layer.

We also explore the effects of intervention layer on three datasets in Figure~\ref{fig:layer_analysis}(a).
For the TweetEmo dataset, the optimal intervention layer is 3. For Amazon-R and Amazon-C datasets, the optimal intervention layers are 7 and 4, respectively.
Notably, intervening at the 5-th or 6-th layer generally improves performance across all three datasets.

\begin{table}[t] \small
    \centering
    \setlength{\tabcolsep}{5pt}
    \begin{tabular}{lcccc}
        \toprule
        \textbf{Position} & \textbf{I.STSB} & \textbf{TweetEmo} & \textbf{Amazon-R}\\
        \midrule
        \textbf{Head} &39.9/37.7   \up{6.0/8.2} &45.5   \up{5.6}&37.0  \up{2.2}\\
        \textbf{FFN} &40.8/39.1  \up{6.9/9.6} &44.2 \up{4.3} &35.7 \up{0.9}\\
        \textbf{Hidden} &39.1/37.2  \up{5.2/7.7} &46.2 \up{6.3} & 35.1 \up{0.3}\\
        \bottomrule
    \end{tabular}
    \caption{Effects of intervention position on three datasets. FFN denotes the output of FFN block, and Hidden denotes the output of the Transformer layer.}
    \label{tab:position}
\end{table}

\textbf{Effects of Scaling Factor}
We investigate the effects of the scaling factor in LLaMA3-8B-Inst using three datasets, as shown in Figure \ref{fig:layer_analysis}(b).
For the TweetEmo dataset, the optimal scaling factor is 15. For Amazon-R and Amazon-C datasets, the optimal scaling factors are 10 and 5, respectively.
Compared with PonTE, most scaling factors can achieve performance improvements. When the scaling factor is 7, 10, or 12, all three datasets perform well.

\begin{table*}[t] \scriptsize \setlength{\tabcolsep}{12pt}
\centering
\begin{tabular}{lrccccc}
\toprule
\textbf{Method} & \textbf{Backbone} & \textbf{I.STSB} & \textbf{TweetEmo} & \textbf{Amazon-R} & \textbf{Mean}\\
\midrule
PonTE & LLaMA3-8B &10.2/9.5 &26.5 &23.4  & 17.4\\
PonTE + SCS-DE (\textbf{\textit{Ours}})& LLaMA3-8B & \cellcolor{tabcolor3} 25.9/24.2 \up{15.7/14.7} & \cellcolor{tabcolor3} 34.8 \up{8.3} &26.9 \up{3.5} \cellcolor{tabcolor3} &  \cellcolor{tabcolor3} 28.0 \up{10.6}\\
\midrule
PonTE & LLaMA3-8B-Inst &33.9/29.5 &39.9 &34.8 &34.5 \\
PonTE + SCS-DE (\textbf{\textit{Ours}}) & LLaMA3-8B-Inst & \cellcolor{tabcolor3} 39.9/37.7 \up{6.0/8.2} & \cellcolor{tabcolor3} 45.5 \up{5.6} & \cellcolor{tabcolor3} 37.0 \up{2.2} & \cellcolor{tabcolor3} 40.0 \up{5.5} \\
\midrule
PonTE & LLaMA3.1-8B &11.3/11.0 &26.2 &23.1 &17.9 \\
PonTE + SCS-DE (\textbf{\textit{Ours}}) & LLaMA3.1-8B & \cellcolor{tabcolor3} 25.5/23.1 \up{14.2/12.1} & \cellcolor{tabcolor3} 38.7 \up{12.5} &25.9 \up{2.8} \cellcolor{tabcolor3} & \cellcolor{tabcolor3} 28.3 \up{10.4} \\
\midrule
PonTE & LLaMA3.1-8B-Inst &29.1/24.2 &40.9  &34.0 &32.1 \\
PonTE + SCS-DE (\textbf{\textit{Ours}}) & LLaMA3.1-8B-Inst & \cellcolor{tabcolor3} 35.9/32.7 \up{6.8/8.5} & \cellcolor{tabcolor3} 44.8 \up{3.9} &36.0 \up{2.0} \cellcolor{tabcolor3} & \cellcolor{tabcolor3} 37.4 \up{5.3}\\
\midrule
PonTE & Qwen2.5-7B & 9.3/9.1 & 35.2 & 25.9  &19.9 \\ 
PonTE + SCS-DE (\textbf{\textit{Ours}}) & Qwen2.5-7B & \cellcolor{tabcolor3} 14.5/11.8 \up{5.2/2.7} & \cellcolor{tabcolor3} 37.7 \up{2.5} & \cellcolor{tabcolor3} 28.2 \up{2.3} & \cellcolor{tabcolor3} 23.1 \up{3.2}\\
\midrule
PonTE & Qwen2.5-7B-Instruct &  27.0/26.3 & 37.2 &  31.6  &30.5 \\
PonTE + SCS-DE (\textbf{\textit{Ours}}) & Qwen2.5-7B-Instruct & \cellcolor{tabcolor3} 32.1/31.0 \up{5.1/4.7} & \cellcolor{tabcolor3} 42.1 \up{4.9} & \cellcolor{tabcolor3} 32.7 \up{1.1} & \cellcolor{tabcolor3} 34.5 \up{4.0}\\
\bottomrule
\end{tabular}
\caption{Results on three datasets using different backbones.}
\label{tab:backbone}
\end{table*}

\begin{table*}[th] \small
    \centering
    \setlength{\tabcolsep}{6pt}
    \begin{tabular}{l ccc}
        \toprule
        \textbf{Prompt} & \textbf{I.STSB} & \textbf{TweetEmo} & \textbf{Amazon-R}\\
        \midrule
        Express this text ``\texttt{[TEXT]}'' in one word in terms of {condition}: `` &  39.9/37.7  \up{6.0/8.2} &45.5 \up{5.6} &37.0 \up{6.2} \\
        Express this text ``\texttt{[TEXT]}'' in one word with respect to {condition}: `` &39.8/38.0 \up{6.8/8.3} &44.6 \up{1.9} &35.8 \up{4.5} \\
        Express this text ``\texttt{[TEXT]}'' in terms of {condition} in one word: `` &37.1/33.1  \up{7.8/8.9} &44.2 \up{2.2} &37.2 \up{0.4} \\
        \midrule
        This text ``\texttt{[TEXT]}'' means in one word in terms of {condition}: `` &36.5/31.5 \up{6.1/7.3} &42.5 \up{0.8}  &37.6 \up{0.9} \\
        This text ``\texttt{[TEXT]}'' means in one word with respect to {condition}: `` &  31.6/28.3 \up{5.0/6.8} &42.8 \up{2.6} &37.2 \up{0.4}\\
        This text ``\texttt{[TEXT]}'' means in terms of {condition} in one word: `` &34.1/29.2 \up{4.2/5.8} &43.2 \up{1.6} &37.7 \up{0.1} \\
        \bottomrule
    \end{tabular}
    \caption{Generalization across different prompts using LLaMA3-8B-Inst as the backbone on three datasets.}
    \label{tab:prompt}
\end{table*}

\begin{table}[t] \small
    \centering
    \setlength{\tabcolsep}{8pt}
    \begin{tabular}{lcc}
        \toprule
        \textbf{Dataset} & \textbf{Condition} & \textbf{V-measure} \\
        \midrule
        \multirow{3}{*}{\textbf{TweetEmo}} 
        & \textbf{the emotion}  & 44.8 \up{3.9}  \\
        & \textbf{the feeling} &42.4 \up{1.2}  \\
        & \textbf{the sentiment} &43.5 \up{4.0}  \\
        \midrule
        \multirow{3}{*}{\textbf{Amazon-R}}
        & \textbf{the star} &32.4 \up{4.7} \\
        & \textbf{the rating}  &35.7 \up{5.0}   \\
        & \textbf{the star rating}  &37.0 \up{2.2}   \\
        \bottomrule
    \end{tabular}
    \caption{Generalization across different conditions using LLaMA3-8B-Inst on two datasets.}
    \label{tab:condition}
\end{table}

\textbf{Effects of Output Layers}
We investigate the effects of output layers in LLaMA3-8B-Inst on two datasets, as shown in Figure \ref{fig:layer_analysis}(c).
SCS-DE consistently improves PonTE on two datasets.
This suggests that our method can enhance the embeddings of all middle and later layers in the LLM, rather than just a single layer.
In particular, our method achieves more significant improvements across all later layers on the Amazon-R dataset.
In particular, our method changes the optimal layer for conditional text embeddings. On the two datasets, PonTE achieves its best performance at layers 30 and 29, respectively, whereas our method reaches the optimum at layers 32 and 28.

\subsection{Generalization Analysis}
\paragraph{Generalization across Different LLMs}
Table \ref{tab:backbone} presents the results across various LLMs, including LLaMA3-8B~\cite{dubey2024llama}, LLaMA3.1-8B~\cite{dubey2024llama}, Qwen2.5-7B~\cite{qwen2.5}, and their instruction-tuned versions.

The results demonstrate that our method achieves consistent improvement across various LLMs, highlighting its generalizability.
Furthermore, instruction-tuned LLMs generally outperform their non-instruction-tuned counterparts, as our task relies more heavily on following the conditions specified in the prompt.
Our method generally achieves greater improvements on LLMs that have not been instruction-tuned, due to their limited instruction-following capability.
Additionally, LLaMA3.1-8B does not outperform LLaMA3-8B in terms of conditional text embeddings.

\paragraph{Generalization across Different Prompts}
Table \ref{tab:prompt} presents the results across various prompts.
Our method achieves improvements across six prompts and reduces variance across prompts.
This demonstrates the generalizability of our method to different prompts.
In addition, we observe that subtle variations in the prompts can have a significant impact on performance, and the optimal prompts differ across datasets.

\paragraph{Generalization across Different Conditions}
Table \ref{tab:condition} presents the results across various conditions on two datasets.
Our method also achieves consistent improvement across 6 conditions.
We can observe that accurate and comprehensive conditions generally lead to better performance.
Moreover, the prompts and conditions that were originally optimal are not necessarily the best after steering.

\section{Conclusion}
In this paper, we introduce a plug-and-play, inference-time self-contrastive steering method to extract conditional text embeddings from LLMs without the need for training or additional data.
SCS creates multi-view embeddings for a given text by modifying the positional encoding and attention mask.
Subsequently, we propose two alternative self-contrastive activation steering strategies to refine the conditional text embeddings.
Extensive experiments on three tasks with their seven associated datasets show that our method can seamlessly improve the performance of existing prompt-based methods across different LLMs in a training-free and plug-and-play manner.

\section*{Limitations}
First, our method requires intervening in the internals of LLMs and is therefore not applicable to black-box LLMs.
Second, we do not validate our method on larger LLMs, such as 70B models.

\section*{Acknowledgments}
We would like to thank the anonymous reviewers for their insightful comments.
This work is supported by the JiangSu Natural Science Foundation under Grant No. BK20251989; the National Natural Science Foundation of China under Grants Nos. 62172208, 62441225, 61972192; the Fundamental and Interdisciplinary Disciplines Breakthrough Plan of the Ministry of Education of China (No.JYB2025XDXM118).
This work is partially supported by Collaborative Innovation Center of Novel Software Technology and Industrialization.

\bibliography{custom}

\appendix

\section{Prompts of Our Method} \label{app:prompt}
Since the conditions for I.STSB and C-STS are specified by the datasets, we report the detailed prompts for the remaining datasets as follows:
\begin{tcolorbox}
\textbf{TweetEmo}: 
Express this text ``\texttt{[TEXT]}'' in one word in terms of \texttt{the emotion}: ``

\textbf{Amazon-R}: 
Express this text ``\texttt{[TEXT]}'' in one word in terms of \texttt{the star rating}: ``

\textbf{Amazon-C}: 
Express this text ``\texttt{[TEXT]}'' in one word in terms of \texttt{the product category}: ``

\textbf{Toxic}: 
Express this text ``\texttt{[TEXT]}'' in one word in terms of \texttt{toxicity}: ``

\textbf{AGNews}: 
Express this text ``\texttt{[TEXT]}'' in one word in terms of \texttt{the topic category of this news}: ``

\end{tcolorbox}

\section{Baselines} \label{app:baseline}
We report the detailed prompts for the baselines as follows:
\begin{tcolorbox}
\textbf{PromptEOL:}  
This sentence: ``\texttt{[TEXT]}'' means in one word: ``

\textbf{PonTE}: 
Express this text ``\texttt{[TEXT]}'' in one word in terms of \texttt{[Condition]}: ``

\textbf{PonTE+Echo}: 
Express this text ``\texttt{[TEXT]} \texttt{[TEXT]}'' in one word in terms of \texttt{[Condition]}: ``

\end{tcolorbox}

\section{Results of Multi-layer Intervention}
Table 8 presents the results of our method under multi-layer intervention.

We observe that single-layer intervention often achieves the best or second-best performance, indicating that intervening on only one layer is sufficient to yield strong performance improvements.
In addition, single-layer intervention involves only one hyperparameter and introduces much lower additional time overhead.

\section{Similarity Measurement Before and After Steering}
We first measure the embedding similarity between PromptEOL and PonTE at layer $\ell$ in Table \ref{tab:cos}, i.e., the cosine similarity of contextualized value vectors before steering.
We observe that even after introducing conditions, the cosine similarity of contextualized value vectors obtained under the two prompts remains very high.
This indicates that PonTE still encodes a substantial amount of general semantic information.

After steering, we measure the embedding similarity again and observe a clear decrease.
This demonstrates that our method can effectively remove general information from the embeddings, making them more focused on the condition.

\section{Results on Other Languages}
Table \ref{tab:other_language} shows the performance of our method on other languages in the Amazon-R dataset.
Our method still achieves performance gains in German and Spanish, demonstrating its cross-lingual generalization capability.

\section{Results on Larger LLMs}
Table \ref{tab:qwen2.5-14b} presents the performance of our method on larger LLMs.

Our method also achieves performance improvements on larger LLMs, which further demonstrates its generalization ability.
Notably, for conditional text embeddings, larger LLMs do not always yield better performance, a finding that has also been observed in previous studies~\cite{lei2024meta,DBLP:conf/acl/FuCJW0L025}.

\section{Case Study}
Table~\ref{tab:case} presents the top-8 tokens decoded by different methods.

The tokens decoded by PonTE include some information that is irrelevant to the instruction, such as ``I'' and ``Family'' in the first text, and ``Rows'' and ``Don'' in the second text.
The tokens decoded by our method are all relevant to the instruction, and their probability distribution is more concentrated.

\begin{table}[t]
    \centering
    \setlength{\tabcolsep}{2pt}
    \begin{tabular}{lcc}
        \toprule
        \textbf{Method} & \textbf{German} & \textbf{Spanish} \\
        \midrule
        \textbf{PonTE}  & 33.0 & 29.0  \\
        \textbf{\textbf{PonTE + SCS-DE (\textbf{\textit{Ours}})} } & 35.8 \up{2.8}  & 31.6 \up{2.6} \\
        \bottomrule
    \end{tabular}
    \caption{Results on the two languages of the Amazon-R dataset.}
    \label{tab:other_language}
\end{table}

\begin{table}[t]
    \centering
    \setlength{\tabcolsep}{4pt}
    \begin{tabular}{ccc}
        \toprule
        \textbf{Dataset} & \textbf{Before Steering} & \textbf{After Steering} \\
        \midrule
        I.STSB  & 0.95 & -0.17  \\
        TweetEmo  & 0.94 & -0.52  \\
        Amazon-R & 0.96    & -0.27        \\
        \bottomrule
    \end{tabular}
    \caption{Similarity before and after steering on three datasets.}
    \label{tab:cos}
\end{table}

\begin{table}[t]
    \centering
    \setlength{\tabcolsep}{8pt}
    \begin{tabular}{ccc}
        \toprule
        \textbf{Layer Index} & \textbf{TweetEmo} & \textbf{Amazon-R} \\
        \midrule
        \multicolumn{3}{l}{\it{Intervening on a single layer:}}\\
        1  & 43.6 & 34.5  \\
        2  & 43.9 & 33.8  \\
        3 & \textbf{45.5}             & 35.4        \\
        4 & 43.3              & 35.7       \\
        5 & 43.9           & 36.5        \\
        6 & 43.9             & 35.8         \\
        7 & 42.7             & \textbf{37.0}        \\
        8 & 44.9             & 36.1       \\
        \midrule
        \multicolumn{3}{l}{\it{Intervening on two layers:}}\\
        1-2 & 41.9 & 31.9 \\
        2-3 & \textbf{44.0} & 35.1 \\
        3-4 & 43.3 & 36.3 \\
        4-5 & 43.3 & 35.9 \\
        5-6 & 43.8 & \textbf{37.0} \\
        6-7 & 42.7 & 36.2 \\
        7-8 & 42.9 & 36.9 \\
        \midrule
        \multicolumn{3}{l}{\it{Intervening on three layers:}}\\
        1-3 & 44.0 & 33.3 \\
        2-4 & \textbf{44.3} & 34.6 \\
        3-5 & 43.0 & 35.4 \\
        4-6 & 43.5 & 36.4 \\
        5-7 & 43.7 & 36.5 \\
        6-8 & 43.3 & \textbf{36.8} \\
        \midrule
        \multicolumn{3}{l}{\it{Intervening on four layers:}}\\
        1-4 & 42.4 & 33.0 \\
        2-5 & 44.4 & \textbf{37.0} \\
        3-6 & 43.1 & 35.4 \\
        4-7 & \textbf{44.9} & 36.3 \\
        5-8 & 43.8 & 36.8 \\
        \midrule
        \multicolumn{3}{l}{\it{Intervening on five layers:}}\\
        1-5 & 44.3 & 31.7 \\
        2-6 & \textbf{45.8} & 34.3 \\
        3-7 & 43.3 & 36.2 \\
        4-8 & 43.6 & \textbf{36.8} \\
        \bottomrule
    \end{tabular}
    \caption{Results of applying multi-layer interventions on the two datasets.}
    \label{tab:multiple_intervention}
\end{table}

\begin{table}[t] \small
\resizebox{\linewidth}{!}{%
\centering
\begin{tabular}{lrcc}
\toprule
\textbf{Method} & \textbf{Backbone} & \textbf{I.STSB} & \textbf{TweetEmo} \\
\midrule
PonTE & Qwen2.5-14B & 13.5/10.9 & 17.4  \\ 
PonTE + SCS-DE (\textbf{\textit{Ours}}) & Qwen2.5-14B & \cellcolor{tabcolor3} 20.2/16.2 \up{6.6/5.3} & \cellcolor{tabcolor3} 27.0 \up{9.6} \\
\midrule
PonTE & Qwen2.5-14B-Instruct &  24.7/21.3 & 22.5  \\
PonTE + SCS-DE (\textbf{\textit{Ours}}) & Qwen2.5-14B-Instruct & \cellcolor{tabcolor3} 28.9/27.5 \up{4.2/5.2} & \cellcolor{tabcolor3} 43.1 \up{20.6}\\
\bottomrule
\end{tabular}
}
\caption{Results on three datasets using different backbones.} \label{tab:qwen2.5-14b}
\end{table}

\begin{table*}[t]
\centering
\small
\setlength{\tabcolsep}{8pt} 
\renewcommand{\arraystretch}{1.5} 
\resizebox{\linewidth}{!}
{%
\begin{tabular}{c c c l} 
\toprule
\textbf{Text} & \textbf{Condition} & \textbf{Method} & \textbf{Top-predicted Tokens and Probability} \\
\midrule
\multirow{4}{*}{\begin{tabular}{@{}c@{}}A man and two children lounge\\ in the living area by a sliding glass door. \end{tabular}}
& \multirow{4}{*}{\begin{tabular}{@{}c@{}}the number of people \end{tabular}}
& \multirow{2}{*}{PonTE}  & Three (0.482), Four (0.054), three (0.046), I (0.038),\\
& & &  TH (0.032), Family (0.03), T (0.029), Two (0.029) \\
\cline{3-4}
& & \multirow{2}{*}{PonTE+SCS-DE} & Three (0.550), three (0.056), Four (0.049), Two (0.037) \\
& & &  TH (0.028), 3 (0.028), T (0.026), Tri (0.024) \\
\midrule
\multirow{4}{*}{\begin{tabular}{@{}c@{}}A row of green donuts, a row of\\ glazed donuts, and a row of chocolate donuts. \end{tabular}}
& \multirow{4}{*}{\begin{tabular}{@{}c@{}}the rows \end{tabular}}
& \multirow{2}{*}{PonTE}  & Rows (0.131), Tr (0.107), Tri (0.105), tr (0.067),\\
& & & Don (0.056), T (0.052), rows (0.05), TR (0.043) \\
\cline{3-4}
& & \multirow{2}{*}{PonTE+SCS-DE} & Three (0.423), Tri (0.073), three (0.070), Rows (0.068)\\
& & & TR (0.038), Tr (0.033), TH (0.032), tr (0.025) \\
\bottomrule
\end{tabular}}
\caption{Top-8 tokens predicted by different methods using LLaMA3-8B-Inst.}\label{tab:case}
\end{table*}

\end{document}